\documentclass{article}
\usepackage{graphicx,mlspconf}
\usepackage{amssymb,amsfonts,amsmath}
\usepackage{mathtools}
\usepackage[linesnumbered,ruled,vlined]{algorithm2e}
\usepackage{xcolor}
\usepackage{soul}
\usepackage{hyperref}
\usepackage{booktabs}
\usepackage{tikz}
\usetikzlibrary{shapes,arrows}
\usetikzlibrary{calc}

\hypersetup{
    colorlinks,
    linkcolor={black},
    citecolor={black},
    urlcolor={black}
}

\copyrightnotice{U.S.\ Government work not protected by U.S.\ copyright}

\copyrightnotice{978-1-5386-5477-4/18/\$31.00 {\copyright}2018 Crown}

\copyrightnotice{978-1-5386-5477-4/18/\$31.00 {\copyright}2018 European Union}

\copyrightnotice{978-1-5386-5477-4/18/\$31.00 {\copyright}2018 IEEE}

\toappear{2018 IEEE International Workshop on Machine Learning for Signal Processing, Sept.\ 17--20, 2018, Aalborg, Denmark}

\newcommand{\w}{\mathbf{w}}
\newcommand{\smlv}{\mathbf{v}}
\newcommand{\smlh}{\mathbf{h}}
\newcommand{\x}{\mathbf{x}}
\newcommand{\z}{\mathbf{z}}
\newcommand{\y}{\mathbf{y}}
\newcommand{\Bigh}{\mathbf{H}}

\newcommand{\R}{{\mathbb{R}}}
\newcommand{\N}{{\mathbb{N}}}

\newcommand{\half}{{\textstyle \frac{1}{2}}}
\newcommand{\vectornorm}[1]{\left|\left|#1\right|\right|}
\newcommand{\ip}[2]{\langle #1,#2 \rangle}

\newcommand{\cvec}{\mathbf{c}}

\newcommand{\eye}{\mathbf{I}}

\title{Scalable Convolutional Dictionary Learning with Constrained Recurrent Sparse Auto-encoders}
\name{Bahareh Tolooshams$^1$\hskip 1in Sourav Dey$^2$\hskip 1inDemba Ba$^1$\thanks{We would like to thank Amazon for their generous support.}}
\address{$^1$School of Engineering and Applied Sciences, Harvard University, Cambridge, MA\\ $^2$Manifold AI, Oakland, CA}
	
\begin{document}
%

\maketitle
\begin{abstract}
Given a convolutional dictionary underlying a set of observed signals, can a carefully designed auto-encoder recover the dictionary in the presence of noise? We introduce an auto-encoder architecture, termed constrained recurrent sparse auto-encoder (CRsAE), that answers this question in the affirmative. Given an input signal and an approximate dictionary, the encoder finds a sparse approximation using FISTA. The decoder reconstructs the signal by applying the dictionary to the output of the encoder. The encoder and decoder in CRsAE parallel the sparse-coding and dictionary update steps in optimization-based alternating-minimization schemes for dictionary learning. As such, the parameters of the encoder and decoder are not independent, a constraint which we enforce for the first time.  We derive the back-propagation algorithm for CRsAE. CRsAE is a framework for blind source separation that, only knowing the number of sources (dictionary elements), and assuming sparsely-many can overlap, is able to separate them. We demonstrate its utility in the context of spike sorting, a source separation problem in computational neuroscience. We demonstrate the ability of CRsAE to recover the underlying dictionary and characterize its sensitivity as a function of SNR. 
\end{abstract}
\section{Introduction}
\label{sec:intro}

In Signal processing, dictionary learning is the de-facto method to learn representations from data. KSVD~\cite{Aharon2006KSVDAA}, alternating-minimization~\cite{Agarwal2016LearningSU} algorithms, and Non-Negative Matrix Factorization (NMF) are all examples of such learning. Here, our focus is on convolutional dictionary learning (CDL), namely when the dictionary of interest has a block-Toeplitz structure induced by a finite number of convolution filters~\cite{GarciaCardona2017ConvolutionalDL}. The authors in~\cite{GarciaCardona2017ConvolutionalDL} provide a thorough review of the state-of-the-art in CDL. Dictionary learning is a bi-convex optimization problem.  Given a set of training examples, existing approaches exploit this structure by alternating between a convex sparse-coding (approximation) step and a convex dictionary-update step, until a convergence criterion is met. While the sparse-coding step, termed convolutional sparse coding (CSC), is embarrassingly parallelizable, the infrastructure to make it seamlessly scalable to thousands of examples has yet to be developed. The most popular approaches to CSC use variants of ADMM~\cite{Boyd2011DistributedOA}. While efficient, ADMM is an iterative optimization method that, at present, lacks the infrastructure that would enable it to be deployed at scale, and to exploit the parallelism offered by GPUs~\cite{sreter2017learned}.
Recent work has explored the connection between CDL and auto-encoders in deep learning~\cite{sreter2017learned,Venkataramani2017NeuralNA}. Framing dictionary learning as training of a neural network opens the possibility of exploiting the GPU-based infrastructure that has been developed for training neural networks, to reduce computational requirements and decrease runtime. In this framework, the \emph{encoder} imitates the sparse coding step in dictionary learning, and the \emph{decoder} reconstructs the input signal from the approximated sparse code, similar to the dictionary-update step. In~\cite{sreter2017learned}, the authors propose an auto-encoder to learn a convolutional dictionary and apply it to image inpainting and denoising. The architecture fails to enforce the constraints that relate the parameters of encoder and decoder. In the absence of such constraints, the parameters lack the interpretability of the convolutional filters in CDL. In addition, the focus in~\cite{sreter2017learned} is on minimization of a given error metric between the input of the auto-encoder and its output, not on the ability of the architecture to learn an underlying convolutional dictionary.



We propose a constrained recurrent sparse auto-encoder architecture (CRsAE) that a) learns true underlying convolutional filters, b) is robust to noise, c) produces sparse codes and performs source separation and d) is substantively faster than approaches based on convex optimization and greedy methods. The encoder in CRsAE is an unrolled recurrent network~\cite{gregor2010learning,Rolfe2013DiscriminativeRS}. We derive the back-propagation for CRsAE and demonstrate its utility in the context of neural spike sorting, a source separation problem in neuroscience in which the sources represent action potentials from different neurons.

The remainder of our treatment begins in Section~\ref{sec:csc} where we introduce our notation and the CDL problem. In Section~\ref{sec:csrae}, we introduce the CRsAE architecture and the accompanying learning algorithm. We show that CRsAE can learn an underlying true dictionary in Section~\ref{sec:recoveryapp} and apply it to neural spike sorting. We conclude in Section~\ref{sec:conc}. 

\vspace*{-2mm}
\section{Convolutional dictionary learning}
\vspace*{-2mm}
\label{sec:csc}
\subsection{Convolutional Generative Model}
\noindent \textbf{Continuous-time model.} Consider a set $(h_c(t))_{c=1}^C$ of $C \in \N^+$ continuous-time filters. Let $t \in [0,T_0)$ and $y_t$ be the continuous-time signal obtained as follows
\vspace*{-2mm}
\begin{equation}\label{eq:CTconv}
y_t = \sum_{c=1}^C \sum_{i=1}^{N_c} x_{c,i}h_c(t-\tau_{ci}) + v_t \quad\textrm{for }t\in [0,T_0)
\end{equation}
\noindent where $v_t$ is i.i.d Gaussian noise with variance $\sigma^2$, $N_c \in \N$ is the number of times $h_c$ appears in the signal, $x_{c,i} \in \R$, and $\tau_{ci} \in \R^+$. This continuous-time observation model applies in a number of signal processing applications where the filters represent sources of interest. Here, we assume that the filters are localized in time, i.e. with support much smaller than $T_0$. This is plausible, for instance, when the filters represent extracellular action potentials, i.e. ``spikes", from a group of neurons observed through an extracellular electrode~\cite{lewicki1998review}. 


\noindent \textbf{Discrete-time model.} In practice, we have access to a discrete-time version $y_n$ of $y_t$ acquired at a sampling rate $f_s$. This leads to the following discrete-time observation model.
\begin{equation}
\label{eq:DTconv}
\begin{aligned}
y_n = \sum_{c=1}^C \sum_{i=1}^{N_c} x_{c,i}h_c[n-n_{ci}] + v_n = \sum_{c=1}^C h_c[n] \ast x_c[n] + v_n\\
\text{for } n = 1,\cdots,\left\lfloor \frac{T_0}{f_s}\right \rfloor, n_{c,i} = \left \lfloor \frac{\tau_{c,i}}{f_s}\right \rfloor
\end{aligned}
\end{equation}
\noindent where, for $c=1,\cdots,C$, $x_c[n] = \sum_{i=1}^{N_c} x_{c,i} \delta [n-n_{c,i}]$ and $(h_c[n])_{n=0}^{K-1}$ is the discrete-time analog of $h_c(t)$. This discrete-time model has attracted a great deal of attention in the signal processing literature, particularly in the dictionary learning literature~\cite{GarciaCardona2017ConvolutionalDL}. In most applications of interest, $K << N = \left\lfloor \frac{T_0}{f_s}\right \rfloor$, owing to the locality of the filters, such as action potentials in the spike-sorting context~\cite{lewicki1998review}. 

\noindent \textbf{Linear-algebraic formulation of discrete-time model.} For $c=1,\cdots,C$, let $\x_c = (x_c[0],\cdots,x_c[N-K])^{\text{T}}$, $\smlh_c = (h_c[0],\cdots,h_c[K-1])^{\text{T}}$, and $N_e = N-K+1$ be the length of each $\x_c$. Let $\Bigh_c \in \R^{N \times N_e}$ be the matrix such that $\Bigh_{c,r,\ell} = h_c[r-1-\ell-1], r=1,\cdots,N; \ell=1,\cdots,N_e$. Stated otherwise, the columns of $\Bigh_c$ consists of delayed versions of the filter $h_c$. With this notation, we can rewrite Eq.~\ref{eq:DTconv} as follows
\vspace*{-2mm}
\begin{equation}
\label{eq:matconv}
\y = \begin{bmatrix} \Bigh_1 | \cdots | \Bigh_C \end{bmatrix} \begin{bmatrix} \x_1 \\ \vdots \\ \x_C \end{bmatrix} + \smlv = \Bigh \x + \smlv
\end{equation}
\noindent Note that, for each $c=1,\cdots,C$, $\x_c$ is an $N_c$-sparse vector with support $\{n_{c,i}+1\}_{i=1}^{N_c}$. This assumption is common in dictionary learning \cite{GarciaCardona2017ConvolutionalDL}. In the spike-sorting setting, we can justify this because the rate of neural spiking is small compared to the length of the recording. \emph{In addition, we will require that only a small number of neurons emit action potentials at the same time, a key condition for the ability to separate the sources of interest}~\cite{Agarwal2016LearningSU}. 

In practice, the data $\y$ are  usually divided into (e.g. non-overlapping) windows of a given size $W$ such that $K << W << N$. In what follows, we assume that Eq.~\ref{eq:matconv} applies locally to each window, i.e.
\begin{equation}\label{eq:conv}
\y_j = \Bigh \x_j + \smlv_j \quad \textrm{for } j = 1,\cdots,J.
\end{equation}
\noindent \textbf{Remark}: The linear-algebraic formulation is primarily for notational convenience. In practice, the matrices involved are large and never stored explicitly. Below we will refer to $\Bigh$ and the filters $(\smlh_c)_{c=1}^C$ that generate the matrix interchangeably.
\vspace*{-2mm}
\subsection{Convolutional Dictionary Learning}

In CDL, the goal is to estimate $(\smlh_c)_{c=1}^C$ by solving the following optimization problem \cite{GarciaCardona2017ConvolutionalDL}:
\vspace*{-1mm}
\begin{equation}\label{eq:DL}
\begin{aligned}
\min_{ (\x_j)_{j=1}^J,(\smlh_c)_{c=1}^C} &\sum_{j=1}^J \frac{1}{2}\vectornorm{\y_j-\Bigh  \x_j}_2 + \lambda \vectornorm{\x_j}_1\\
\text{ s.t. } &\vectornorm{\smlh_c}_2 \leq 1 \quad \textrm{for } c = 1,\cdots, C,
\end{aligned}
\end{equation}
\noindent where $\lambda > 0$ is a regularization parameter enforcing sparsity. Starting with an initial guess for the filters, the method of choice to solve Eq.~\ref{eq:DL} is to alternate between a CSC step, which solves for the sparse codes $(\x_j)_{j=1}^J$ given the filters, and a step that solves for the filters given the sparse codes. 
\vspace*{-2mm}
\subsubsection{Convolutional Sparse Coding Update}

Given the filters, the CSC problem is the solution to $J$ independent convex optimization problems, i.e. it is separable over $j$. The $j^{\text{th}}$ sparse code $\x_j$ is the solution to
\vspace*{-1mm}
\begin{equation}\label{eq:csc_update}
\min_{\x_j} \frac{1}{2} \vectornorm{\y_j-\Bigh  \x_j}_2 + \lambda \vectornorm{\x_j}_1
\end{equation}
Eq. \ref{eq:csc_update} can be solved efficiently through ADMM, as in \cite{GarciaCardona2017ConvolutionalDL}. At present, ADMM lacks the infrastructure that would enable it to be deployed at scale seamlessly, and to exploit the parallelism offered by GPUs~\cite{sreter2017learned} to solve Eq.~\ref{eq:csc_update} for all $J$ windows simultaneously.

\vspace*{-2mm}
\subsubsection{Dictionary Update}
The updated filters are the solution to
\vspace*{-1mm}
\begin{equation}\label{eq:dict_update}
\min_{(\smlh_c)_{c=1}^C} \sum_{j=1}^J \frac{1}{2} \vectornorm{\y_j-\Bigh  \x_j}_2 \text{ s.t. } \vectornorm{\smlh_c}_2 
\leq 1 \quad \textrm{for } c = 1,\cdots, C.
\end{equation}
\noindent This is a convex optimization problem for which various approaches have been proposed~\cite{GarciaCardona2017ConvolutionalDL}. Irrespective of the approach, the filter update step is computationally expensive and \emph{not} parallelizable over the $J$ training examples. 
\vspace*{-2mm}
\section{Convolutional, recurrent, constrained, sparse auto-encoders}
\label{sec:csrae}
\vspace*{-2mm}
In this section, we introduce an auto-encoder architecture for CDL. This auto-encoder has an implicit connection with the alternating-minimization algorithm for dictionary learning described in the previous section.

Given $\Bigh$, the encoder produces a sparse code using a finite (large) number of iterations of the FISTA algorithm~\cite{beck2009fast}. To reconstruct $\y$, the decoder applies $\Bigh$ to the output of the encoder. We call this architecture a constrained recurrent sparse auto-encoder (CRsAE). The constraint comes from the fact that the operations used by the encoder and decoder are tied to each other through $\Bigh$. Hence, the encoder and decoder are not independent, unlike in previous work~\cite{sreter2017learned}. The encoder from CRsAE is an unrolled recurrent network for which all of the steps share the same inputs~\cite{gregor2010learning,Rolfe2013DiscriminativeRS}. This recurrent behaviour is crucial to producing the sparse codes necessary for successful dictionary learning~\cite{Agarwal2016LearningSU}.


Constraining the encoder and decoder as in CRsAE makes the connection between auto-encoders and dictionary learning more explicit. Indeed, CRsAE filters are directly  interpretable in terms of the model in Eq. \ref{eq:DTconv}. Moreover, CRsAE now has $\frac{1}{3}$ of the number of parameters of the auto-encoder from~\cite{sreter2017learned}. 
\vspace*{-2mm}
\subsection{Forward Propagation: CRsAE encoder}
To simplify notation, and because the gradient of our objective function is separable with respect to $j$, we drop the subscript $j$. 

\noindent \textbf{Remark}: For an arbitrary vector $\z \in \R^N$, the vector $\z_t$ refers to the vector at the $t^{\text{th}}$ iteration of an iterative algorithm, and $z_{t,n}$ to the $n^{\text{th}}$ entry of $\z_t$. This will be made clear when necessary through the specification of the range of the indices.

FISTA is a fast and iterative procedure to find a sparse solution to the noisy linear system from Eq.~\ref{eq:csc_update}. Given the input $\y$, an initial guess $\x_0$, and letting $\x_{-1} = \x_0$, $T$ iterations of FISTA will generate a sequence $(\x_t)_{t=1}^T$ for which each element in the sequence depends on the past two previous elements. The last element $\x_T$ corresponds to output of the encoder. We introduce a ``state'' vector $\z_t = \begin{bmatrix}\z_{t}^{(1)} \\ \z_{t}^{(2)}\end{bmatrix} = \begin{bmatrix}\x_{t}\\ \x_{t-1}\end{bmatrix}$ that will simplify the derivation of the back-propagation algorithm for computing the gradient of the loss function of the auto-encoder with respect to $\smlh = (\smlh_1^\text{T},\cdots,\smlh_C^\text{T})^\text{T}$.

The encoder in CRsAE uses $\Bigh$ and $\Bigh^{\textrm{T}}$ to perform $T$ iterations of FISTA. In our convolutional setting, $\Bigh$ computes a sum of convolutions as in Eq.~\ref{eq:matconv}, and $\Bigh^{\textrm{T}}: \R^{N} \to \R^{N_eC}$ performs correlation between its input and $(\smlh_c)_{c=1}^C$. The detailed steps are given in Algorithm~\ref{algo:encoderfprop}, where $\eta_{\epsilon}: \R^{N_e C} \to \R^{N_e C}$ is the shrinkage operator that applies element-wise shrinkage to its input. The shrinkage operator is defined by
\begin{equation}\label{eq:shrinkage}
\eta_{\epsilon,n}(\z) = (|z_n|-\epsilon)_+ \textrm{sgn}(z_n)
\end{equation}
\vspace*{-5mm}
\begin{algorithm}
\KwIn{$\y, \Bigh, \lambda, L \geq \sigma_\text{max}(\Bigh^{\text{T}}\Bigh)$}
\KwOut{$\z_{T}$}
$\z_0 = \mathbf{0},s_0 = 0$\\
\For{$t =1$ to $T$}{
$s_t = \frac{1 + \sqrt{1+4s_{t-1}^2}}{2}$\\
$\w_t = \z_{t-1}^{(1)} + \frac{s_{t-1}-1}{s_t}\left(\z_{t-1}^{(1)}-\z_{t-1}^{(2)}\right) = \begin{bmatrix}\left(1 + \frac{s_{t-1}-1}{s_t}\right) \eye_{N_e}| - \frac{s_{t-1}-1}{s_t} \eye_{N_e}\end{bmatrix} \z_{t-1}$ \\
$\cvec_t = \w_t + \frac{1}{L} \Bigh^{\text{T}}(\y-\Bigh\w_t)$\\
$\z_t = \begin{bmatrix}\eta_{\frac{\lambda}{L}}(\cvec_t) \\ \z_{t-1}^{(1)}\end{bmatrix}$
}
\caption{Encoder in CRsAE}
\label{algo:encoderfprop}
\end{algorithm}

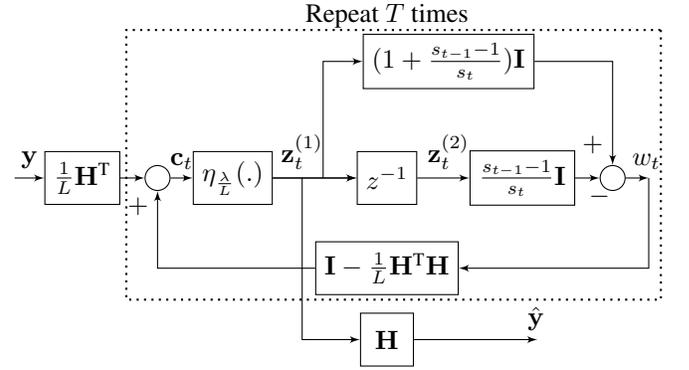
\begin{figure}[htb]
\begin{minipage}[b]{1.0\linewidth}
  \centering
\tikzstyle{block} = [draw, fill=none, rectangle, 
    minimum height=2em, minimum width=2em]
\tikzstyle{sum} = [draw, fill=none, circle, node distance=1cm]
\tikzstyle{input} = [coordinate]
\tikzstyle{output} = [coordinate]
\tikzstyle{pinstyle} = [pin edge={to-,thin,black}]
\begin{tikzpicture}[auto, node distance=2cm,>=latex']
	cloud/.style={
      draw=red,
      thick,
      ellipse,
      fill=none,
      minimum height=1em}
    \node [input, name=input] {};
    \node [block, node distance=0.9cm, right of=input] (HT) {$\frac{1}{L}\Bigh^\textrm{T}$};
    \node [sum, node distance=1cm, right of=HT] (sum) {};
    \node [block, right of=sum, node distance=1.cm] (shrinkage) {$\eta_{\frac{\lambda}{L}}(.)$};
    \draw [->] (sum) -- node[name=u] {$\cvec_t$} (shrinkage);
    \node [block, node distance=2.05cm, right of=shrinkage] (delay) {$z^{-1}$};
    \node [block, node distance=1.8cm, right of=delay] (t) {$\frac{s_{t-1}-1}{s_t}\eye$};
    \node [block, node distance=1.2cm, below of=delay] (feedback) {$\eye-\frac{1}{L}\Bigh^\textrm{T}\Bigh$};
     
    \node [sum, node distance=1.2cm, right of=t] (sub) {};
    \node [output, node distance=0.48cm, right of=sub] (output) {};
  
    \node [rectangle, fill=none, node distance=2.16cm, above of=delay] (text) {Repeat $T$ times};
    \draw [draw,->] (input) -- node {$\y$} (HT);
    \draw [->] (HT) -- node {} (sum);
    \draw [->] (shrinkage) -- node [above,pos=0.35,name=Z] {$\z_t^{(1)}$}(delay);

    \draw [->] (shrinkage) -- node [below,name=temp,pos=0.6] {}(delay);

    \draw [->] (feedback) -| node[pos=0.90] {$+$} 
        node [near end] {} (sum);
    \draw [->] (delay) -- node[name=Z2, pos=0.6] {$\z_t^{(2)}$} (t);
    \node [block, node distance=1.2cm, above of=Z2] (forward) {$(1 + \frac{s_{t-1}-1}{s_t})\eye$};
    \draw [->] (temp) |- (forward);
    \draw [->] (forward) -| node[left,pos=0.90] {$\small{+}$} 
        node [near end] {} (sub);
	\draw [->] (t) -- node[pos=0.99, below] {$-$} (sub);	
	\draw[thick,dotted]     ($(forward.north east)+(+1.68,0.05)$) rectangle ($(HT.south east)+(+0.08,-1.26)$);
	\draw [->] (sub) -- node[name=wt,pos=0.90] {$w_t$} (output);
	\draw [->] (output) |- (feedback);

	\node [block, node distance=0.95cm, below of=feedback] (H) {$\Bigh$};
	\draw [->] (Z) |- (H);
	\node[output, node distance=2cm, right of=H] (yhat) {};
	\draw [->] (H) -- node[pos=0.99] {$\hat \y$} (yhat);
\end{tikzpicture}
\end{minipage}
\caption{Block Diagram of CRsAE. Given a convolutional dictionary, the encoder performs $T$ iterations of FISTA. The decoder applies the convolutional dictionary to the output of the encoder. The operator $z^{-1}$ refers to the delay in discrete-time.}
\label{fig:blockdiagram}
\end{figure}

\noindent The decoder uses the sparse code from the last FISTA iteration and $\Bigh$ to generate a reconstruction of $\y$ as follows:
\begin{equation}\label{eq:decoder}
\hat \y = \Bigh \z_{T}^{(1)} = \begin{bmatrix} \Bigh | \mathbf{0}_{N \times N_eC} \end{bmatrix} \z_{T}
\end{equation}
At last, CRsAE minimizes the following loss function:
\begin{equation}\label{eq:loss}
\mathcal{L}(\y,\hat \y) = \half \vectornorm{\y-\hat \y}_2^2
\end{equation}
The block diagram of CRsAE is represented in Figure \ref{fig:blockdiagram}.

\vspace*{-4mm}

\vspace*{-2mm}
\subsection{CRsAE Gradient Backpropagation}
Estimating the dictionary with the afore-mentioned architecture leads us to the following optimization problem
\vspace*{-2mm}
\begin{equation}\label{eq:backprop_op_2}
\min_{(\smlh_c)_{c=1}^C} \sum_{j=1}^J \mathcal{L}(\y_j;\hat \y_j) \text{ s.t. } \vectornorm{\smlh_c}_2 \leq 1, c = 1,\cdots, C.
\end{equation}
Algorithm \ref{algo:fistabprop} shows the back-propagation of CRsAE. Note that there is only one set of filters $(\smlh_c)_{c=1}^C$ that are shared by all layers of the network. In other words, there are only $K \times C$ trainable parameters in CRsAE, where $K$ is the length of each filter $\smlh_c$ and $C$ is the number of filters. The derivation of the back-propagation is provided in detail in Appendix. The notation $\delta \cdot$, from~\cite{gregor2010learning}, refers to the gradient of the loss function with respect to the variable that follows. We include this derivation for completeness. In practice, we use Tensorflow's autograd functionality to compute the gradient.


\begin{algorithm}
\KwIn{$\y,\lambda,L,\smlh$, Variables $s_t, \w_t, \cvec_t, \z_T$ from forward propagation.}
\KwOut{$\delta \smlh$}
$\delta \hat \y = \hat \y - \y, \delta \smlh = \mathbf{0}_K$\\
$\delta \cvec_{T+1} = \frac{\partial \hat \y}{\partial \cvec_{T+1}} \delta \hat \y $\\
$\delta \smlh = \delta \smlh +  \frac{\partial \cvec_{T+1}}{\partial \smlh} \delta \cvec_{T+1}$\\
$\delta \z_{T} = \frac{\partial \cvec_{T+1}}{\partial \z_{T}} \delta \cvec_{T+1}$\\
\For{$t =T$ to $1$}{
$\delta \cvec_t = \frac{\partial \z_t}{\partial \cvec_t} \delta \z_t $\\
$\delta \smlh = \delta \smlh +  \frac{\partial \cvec_t}{\partial \smlh} \delta \cvec_t$\\
$\delta \z_{t-1} = \frac{\partial \cvec_t}{\partial \z_{t-1}} \delta \cvec_t$
}
\caption{Back propagation for CRsAE}
\label{algo:fistabprop}
\end{algorithm}
\vspace*{-5.5mm}
\section{Applications of CR\lowercase{s}AE to dictionary learning and neural spike sorting}
\label{sec:recoveryapp}
\vspace*{-2mm}
\subsection{Can a sparse recurrent auto-encoder learn a convolutional dictionary?}
\vspace*{-1mm}
To address the question of whether CRsAE can learn a convolutional dictionary, we simulated extracellular voltage recordings of $C=3$ neurons obtained using an array of $4$ electrodes implanted in the brain. We let the correlation $\ip{\smlh_c}{\smlh_{c'}}$ ($c \neq c'$) between filters range between $-0.087$ and $0.455$. For each $c=1,\cdots,3$,  we assumed that the amplitudes $(x_{c,i})_{i=1}^{N_c}$ of the action potentials are i.i.d. $\mathcal{N}(\mu_c,\sigma^2_c)$, $(\mu_1,\sigma_1) = (362,20)$, $(\mu_2,\sigma_2) = (388,25)$ and $(\mu_3,\sigma_3) = (360,30)$. The units of the simulated recordings are in 
$mV$.



\vspace*{-2mm}
\vspace*{-2mm}
\subsubsection{Simulated data}
\vspace*{-1mm}
\textbf{Continuous-time model simulation}. Each simulated recording lasted $T_0 = 18$ $s$. Consistent with the biophysics of neurons~\cite{lewicki1998review}, we picked the filters (action potentials) to last $1.5$ $ms$. Due to the refractory period of neurons, whose length is on the same order as the filter length, we simulated the data such that the action potentials (filters) from the same neuron do not overlap.  We allowed the filters from different neurons to overlap. We assumed each electrode is an independent realization of the model from Eq. \ref{eq:CTconv}. We assumed an average spiking rate of $30$ Hz for each simulated neuron.


\noindent \textbf{Discrete-time model simulation}. Assuming a sampling rate $f_s = 30$ kHz, $\smlh_c \in \R^{45}$ and, for each of the electrodes, $\y \in \R^{540,000}$. Assuming a window length of $0.1$ $s$, $\y_j \in \R^{3000}$, for $j=1,\cdots,720$. With our assumptions on the spiking rate of each neuron, $\x_j$ is on average $3 \times 3 = 9$ sparse.  
\vspace*{-2mm}
\vspace*{-2mm}
\subsection{Training parameters}
\vspace*{-3mm}
For training, validation, and testing, we used $630$, $70$, and $20$ windows respectively. We trained CRsAE using mini-batch gradient descent back-propagation with the ADAM optimizer~\cite{Kingma2014AdamAM} for a maximum of $60$ epochs. In cases when the filters converged early, and the validation loss stopped to improve, we stopped the training. We picked the learned dictionary as the one with minimum validation loss. The number of training parameters was $K \times C = 45 \times 3 = 135$. The parameters to be tuned were $\lambda$, $L$, the learning rate of the ADAM optimizer, the number of FISTA iterations $T$, and the mini-batch size $B$. Below, we discussed how we selected each.

\noindent \textbf{Regularization parameter $\lambda$}. Let $\Bigh_0$ denote the initial dictionary estimate: $\y_j = \Bigh_{0} \x_j + \underbrace{(\Bigh-\Bigh_{0}) \x_j + \smlv_j}_{\mathbf{n}_j}$. The observations contain two sources of noise, namely observation noise, and noise from lack of knowledge of $\Bigh$. For simulated data, we can compute both quantities explicitly. If we let $\hat{\sigma}_n = \frac{1}{N} \vectornorm{n_j}_2$, we can use $\lambda \approx \hat{\sigma}_n \sqrt{2\log( C\times N_e)}$  as suggested in~\cite{Chen1998AtomicDB}. For real data, we can estimate $\vectornorm{\smlv_j}_2$ from ``silent'' periods. Assuming $\Bigh_0$ is close enough, an assumption well-justified by the dictionary learning literature~\cite{Agarwal2016LearningSU}, we propose to estimate the cumulative noise by scaling this estimate by a constant factor of $1.1$ to $2$. In principle, this argument suggests we should decrease $\lambda$ as the estimate of the dictionary improved. The rate at which this occurs should depend on the rate at which the dictionary estimate improves. To our knowledge, there is no theory characterizing this. Therefore, we do not decrease $\lambda$ in the results we report.

\noindent \textbf{Picking $L$ and the learning rate}. $L$ must be greater than the maximum eigenvalue of $\Bigh^\textrm{T}\Bigh$~\cite{beck2009fast}. We used an existing collection of neural action potentials to estimate $L$. We set $L = 26$. We varied the learning rate of the optimizer from $10^{-5}$ to $10^{-1}$ and chose the one associated with the sharpest drop in the validation loss function as in~\cite{Smith2017CyclicalLR}.


\noindent \textbf{Number of FISTA iterations $T$}. The importance of the encoder producing sparse codes, even for approximate dictionaries, is theoretically well-known~\cite{Agarwal2016LearningSU}. Therefore, we picked a large number $T \approx 200$ of FISTA iterations, which was sufficient to produce sparse codes at the output of the encoder. Along with the constraints on the encoder and decoder, we believe this is why CRsAE is successful at CDL where other approaches are not~\cite{sreter2017learned}, as our results will demonstrate. 

\noindent \textbf{Batch size $B$}. We chose $B = 16$. We found this to be a choice for which the ADAM optimizer successfully avoided local optima of the loss function. 

\begin{figure}[htb]
\begin{minipage}[b]{1.0\linewidth}
  \centering
  \centerline{\includegraphics[width=8.5cm]{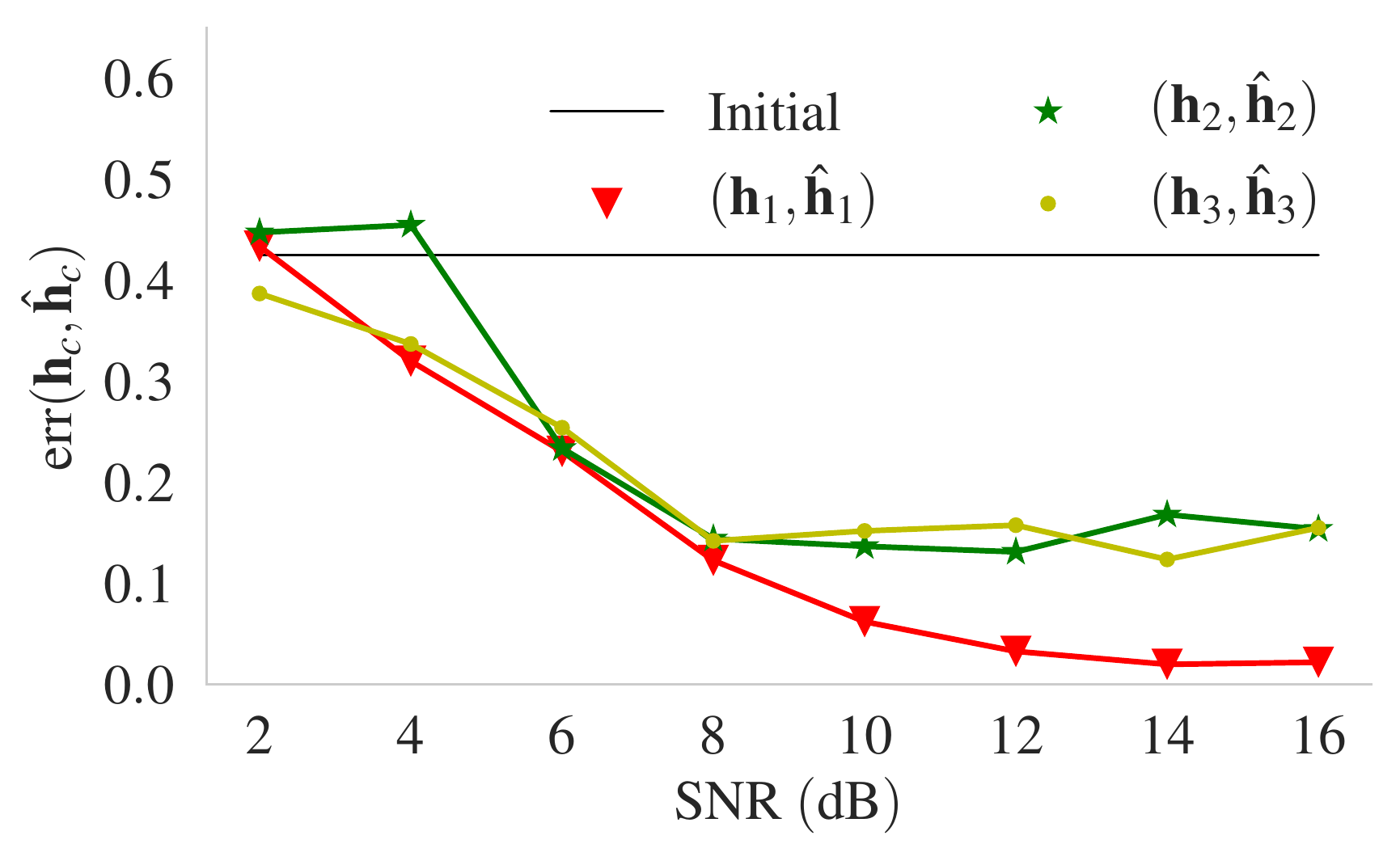}}
\end{minipage}
\caption{Error $\text{err}(\smlh_c,\hat{\smlh}_c)$ from CRsAE as a function of SNR.}
\label{fig:SNR}
\end{figure}

\begin{figure}[htb]
\begin{minipage}[b]{1.0\linewidth}
  \centering
  \centerline{\includegraphics[width=8.5cm]{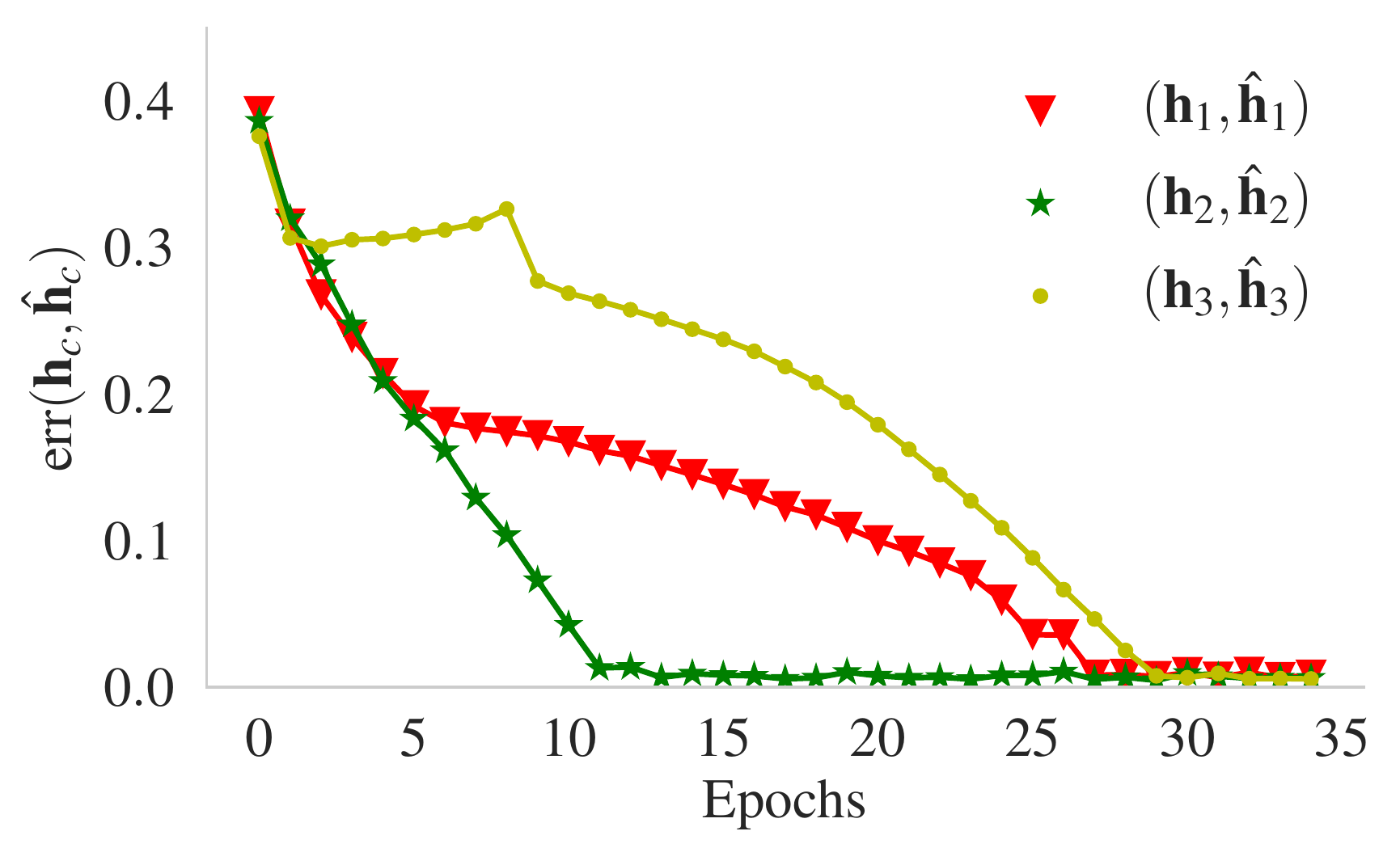}}
\end{minipage}
\caption{Error $\text{err}(\smlh_c,\hat{\smlh}_c)$ from CRsAE as a function of the number of epochs.}
\label{fig:err_dist_epochs}
\end{figure}

\begin{figure}[htb]
\begin{minipage}[b]{1.0\linewidth}
  \centering
  \centerline{\includegraphics[width=8.5cm]{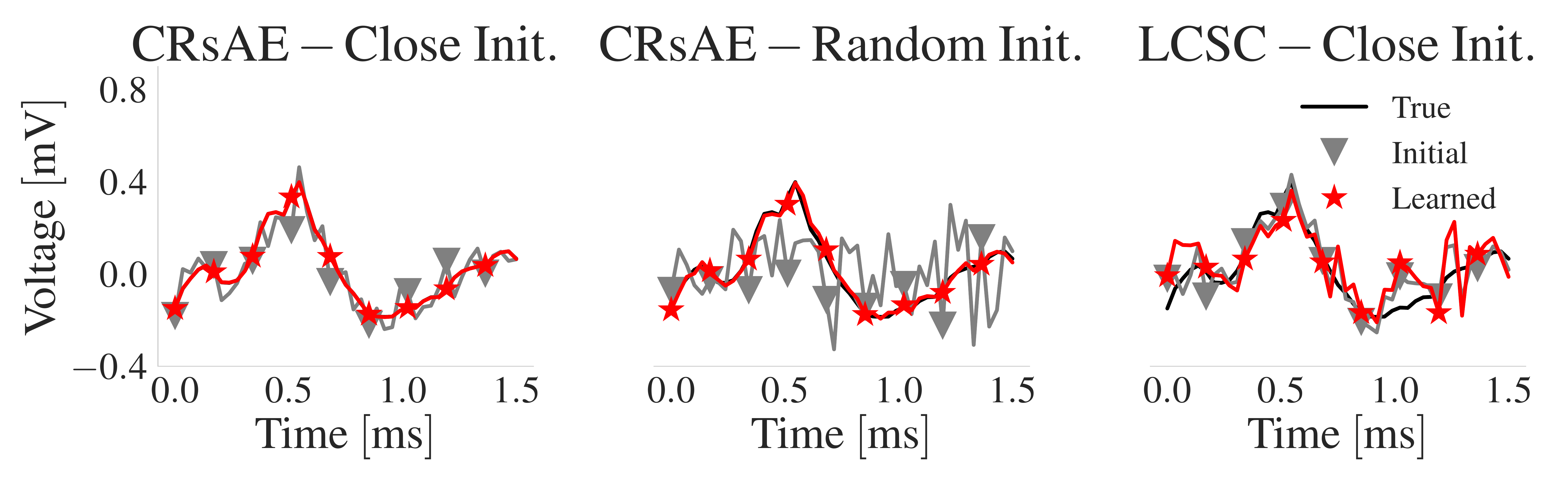}}
\end{minipage}
\caption{Learned Dictionary of CRsAE vs LCSC.}
\label{fig:learned}
\end{figure}

\noindent \textbf{Results.} We used simulated data to assess the ability of CRsAE to recover an underlying convolutional dictionary in the presence of noise. We trained CRsAE using the Keras framework on AWS, with TenorFlow as a backend\footnote{\url{https://github.com/ds2p/crsae}}. Let $\hat{\smlh}_c$ be the estimate from CRsAE of the underlying dictionary. We measured accuracy using the following standard measure~\cite{Agarwal2016LearningSU}
\vspace*{-1mm}
\begin{equation}\label{eq:distance_error}
\text{err}(\hat{\smlh}_c,\smlh_c) = \sqrt{1 - \frac{\ip{\smlh_c}{\hat{\smlh}_c}^2}{\|{\smlh}_c\|_2^2\|\hat{\smlh}_c\|_2^2}}
\end{equation}
The smaller this measure, which ranges from $0$ to $1$, the closer the learned dictionary to the true one.
\noindent Figure \ref{fig:SNR} demonstrates the ability of CRsAE to recover the filters used in the simulation in the presence of noise. 
For each point in Figure~\ref{fig:SNR}, we estimated the recovery error by averaging over $7$ independent simulations. We note that we only used $700$ examples per simulation, which took $\approx 1$ hour on a GPU! Increasing the number of simulations and the number of training examples would let the errors converge to values much closer to $0$.  Figure~\ref{fig:err_dist_epochs} demonstrates that, for one of the best simulations with $16$ dB SNR, the errors do converge to $0$.
\noindent We also compared the dictionary-learning performance of CRsAE to the network in \cite{sreter2017learned} where the encoder consists of $3$ ISTA iterations, and the decoder is linear and \emph{unconstrained}. We term this architecture LCSC(3). Figure~\ref{fig:learned} compares the true dictionary to that estimated using LCSC(3) when the SNR is 16 dB. We considered two cases for CRsAE. In the first, $\Bigh_0$ was a random perturbation of the true dictionary $\Bigh$ such that the error (Eq.~\ref{eq:distance_error}) was between $0.4$ to $0.5$. In the second, the entries of $\Bigh_0$ were assumed to be i.i.d. Gaussian. For LCSC(3), we initialized the dictionary as in the first case for CRsAE. We also allowed $\lambda$ to be trainable as in~\cite{sreter2017learned}. In spite of the fact LCSC(3) converged to a solution with small reconstruction error,  unlike CRsAE, it was not able to learn the true filters.

\vspace*{-4mm}
\section{Conclusion}
\label{sec:conc}
\vspace*{-3mm}
We proposed a constrained, recurrent, sparse, auto-encoder architecture--CRsAE--whose parameters are interpretable in the convolutional dictionary learning setting. In CRsAE, the parameters from the recurrent layers of the encoder are tied to one another, so are they tied to the parameters of the decoder. The small number of trainable parameters enables CRsAE to learn with a relatively small number of examples and training epochs. We demonstrated its ability to reliably estimate action potentials (filters) from $3$ neurons using simulated, 4-channel, extracellular voltage recordings with varying levels of SNR. We compared CRsAE to previous approaches that proposed auto-encoder architectures for convolutional dictionary learning and found that these were unable to learn the underlying filters, despite the small auto-encoder input-output error. We attribute our success to two factors. The first is the explicit enforcement of the constraint on the encoder and decoder parameters. The second is the depth of the encoder in CRsAE: a deep encoder (large number of FISTA iterations) guarantees close-to sparse codes, in the presence of noise and/or imperfect initial filters. In future work, we will generalize CRsAE to dictionary learning for multi-dimensional signals.
%
%
%
\vspace*{-4mm}
\section{Appendix}
\label{sec:app}
\vspace*{-3mm}
We give a derivation of the back-propagation algorithm for CRsAE. For simplicity, we restrict the derivation to the case of a single convolution kernel/filter. Note that we include the derivation for completeness, but in practice, we use Tensorflow's autograd functionality to compute the gradient.

\noindent \textbf{Back propagation}: Let $\smlh = (h[0],h[1],\cdots,h[K-1])^{\text{T}}$. We can assume, without loss of generality, that $J=1$ in Eq.~\ref{eq:loss}. We would like to evaluate the gradient of $\mathcal{L}(\y,\hat \y) = \half \vectornorm{\y-\hat \y}_2^2$ with respect to $\smlh$:
\begin{equation}
	\frac{\partial \mathcal{L}}{\partial \smlh} = \sum_{t=1}^{T+1}\frac{\partial \cvec_t}{\partial \smlh} \frac{\partial \mathcal{L}}{\partial \cvec_t} = \sum_{t=1}^{T+1} \frac{\partial \cvec_t}{\partial \smlh} \delta \cvec_t,
	\label{eq:partialh}
\end{equation}
\noindent where the notation $\delta \cdot$, from~\cite{gregor2010learning}, refers to the gradient of the loss function with respect to the variable that follows.
\vspace*{-1mm}
\begin{equation}
	\delta \cvec_t = \frac{\partial \z_t}{\partial \cvec_t} \frac{\partial \mathcal{L}}{\partial \z_t} = \frac{\partial \z_t}{\partial \cvec_t}  \delta \z_t.
	\label{eq:particalct}
\end{equation}
If we could evaluate Eq.~\ref{eq:particalct} for every $t = 1,\cdots,T$, then we can evaluate the second half of each term of Eq.~\ref{eq:partialh}. We will compute the first half using the equations from Algorithm~\ref{algo:encoderfprop}. We can evaluate Eq.~\ref{eq:particalct} through the following recursion
\begin{equation}
	\delta \z_{t-1} = \frac{\partial \z_t}{\partial \z_{t-1}} \frac{\partial \mathcal{L}}{\partial \z_t} = \frac{\partial \cvec_t}{\partial \z_{t-1}}\frac{\partial \z_t}{\partial \cvec_{t}} \frac{\partial \mathcal{L}}{\partial \z_t} = \frac{\partial \cvec_t}{\partial \z_{t-1}} \delta \cvec_t.
	\label{eq:partialzt1}
\end{equation}
We initialize the recursion with
\begin{equation}
	\delta \cvec_{T+1} = \frac{\partial \mathcal{L}}{\partial \cvec_{T+1}} = \frac{\partial \hat \y}{\partial \cvec_{T+1}} \frac{\partial \mathcal{L}}{\partial \hat \y} = \eye_{N} \delta \hat \y = \hat \y-\y.
	\end{equation}
	
\noindent Finally, before we can evaluate the full gradient propagation procedure, we need to specify $\frac{\partial \cvec_{t}}{\partial \smlh}$ (Eq.~\ref{eq:partialh}), $\frac{\partial \z_t}{\partial \cvec_t}$ (Eq.~\ref{eq:particalct}) and $\frac{\partial \cvec_{t}}{\partial \z_{t-1}}$ (Eq.~\ref{eq:partialzt1}).
\vspace*{-2mm}
\begin{equation}
	\frac{\partial \cvec_{T+1}}{\partial \smlh} = \mathbf{Z}_{T\ast}^{(1)}, \in \R^{K \times N}.
\end{equation}
\noindent where
\vspace*{-2mm}
\begin{equation}
\mathbf{Z}_{T\ast}^{(1)} = \begin{bmatrix}
    \z_{T,0}^{(1)} & \z_{T,1}^{{(1)}} & \z_{T,2}^{(1)} & \dots  & \dots \\
    0 & \z_{T,0}^{(1)} & \z_{T,1}^{(1)} & \dots  & \dots \\
    \vdots & \vdots & \z_{T,0}^{(1)} & \ddots & \vdots \\
    0 & 0 & 0 & \dots  & \dots
\end{bmatrix}
\end{equation}
\noindent and if $t=1,\cdots,T$
\vspace*{-1mm}
\begin{equation}
\left(\frac{\partial \cvec_{t}}{\partial \smlh}\right)_{m,n} = \frac{1}{L}\sum_{\{|k|< K\}} w_{t,n-1+k} \frac{\partial}{\partial h_{m-1}} c_{hh,-k} + y_{m-1+n-1}.
\end{equation}
\noindent where $m = 1,\cdots, K; n = 1,\cdots, N_e$, $ 0 \leq n-1+k \leq N_e-1$, and $c_{hh,k} = h_k \ast h_{-k}$ is the deterministic auto-correlation function of $(h_n)_{n=0}^{K-1}$.
\vspace*{-3mm}
\[
  \frac{\partial \hat \y}{\partial \cvec_{T+1}}=\eye_{N}
\]
\[
  \frac{\partial \z_{t}}{\partial \cvec_{t}}=
               \begin{bmatrix}
               	\text{diag}(\eta_{\frac{\lambda}{L}}'(\cvec_t))| \mathbf{0}_{N_e} 
               \end{bmatrix} \in \R^{N_e \times 2N_e}, t=1,\cdots, T.
\]
\vspace*{-3mm}
\noindent Finally,
\[
  \frac{\partial \cvec_{t}}{\partial \z_{t-1}}=\begin{cases}
                              \begin{bmatrix}
               	\Bigh^{\text{T}} \\ \mathbf{0}_{N_e \times N}
               \end{bmatrix} \in \R^{2N_e \times N}, \text{if } t=T+1\\
               \begin{bmatrix}
               	\left(1 + \frac{s_{t-1}-1}{s_t}\right) \eye_{N_e} \\  -\frac{s_{t-1}-1}{s_t} \eye_{N_e}
               \end{bmatrix}
               \begin{bmatrix}
               	\eye_{N_e} - \frac{1}{L} \Bigh^{\text{T}}\Bigh.
               \end{bmatrix}
               \text{, otherwise}.
            \end{cases}
\]

%

\bibliographystyle{IEEEbib}

\bibliography{mlsp18.bib}

\begin{thebibliography}{10}

\bibitem{Aharon2006KSVDAA}
Michal Aharon, Michael Elad, and Alfred Bruckstein,
\newblock ``K-svd: An algorithm for designing overcomplete dictionaries for
  sparse representation,''
\newblock 2006.

\bibitem{Agarwal2016LearningSU}
Alekh Agarwal, Anima Anandkumar, Prateek Jain, Praneeth Netrapalli, and Rashish
  Tandon,
\newblock ``Learning sparsely used overcomplete dictionaries via alternating
  minimization,''
\newblock {\em SIAM Journal on Optimization}, vol. 26, pp. 2775--2799, 2016.

\bibitem{GarciaCardona2017ConvolutionalDL}
Cristina Garcia-Cardona and Brendt Wohlberg,
\newblock ``Convolutional dictionary learning,''
\newblock {\em CoRR}, vol. abs/1709.02893, 2017.

\bibitem{Boyd2011DistributedOA}
Stephen~P. Boyd, Neal Parikh, Eric Chu, Borja Peleato, and Jonathan Eckstein,
\newblock ``Distributed optimization and statistical learning via the
  alternating direction method of multipliers,''
\newblock {\em Foundations and Trends in Machine Learning}, vol. 3, pp. 1--122,
  2011.

\bibitem{sreter2017learned}
Hillel Sreter and Raja Giryes,
\newblock ``Learned convolutional sparse coding,''
\newblock {\em arXiv preprint arXiv:1711.00328}, 2017.

\bibitem{Venkataramani2017NeuralNA}
Shrikant Venkataramani, Y.~Cem S{\"u}bakan, and Paris Smaragdis,
\newblock ``Neural network alternatives toconvolutive audio models for source
  separation,''
\newblock {\em 2017 IEEE 27th International Workshop on Machine Learning for
  Signal Processing (MLSP)}, pp. 1--6, 2017.

\bibitem{gregor2010learning}
Karol Gregor and Yann LeCun,
\newblock ``Learning fast approximations of sparse coding,''
\newblock in {\em Proceedings of the 27th International Conference on Machine
  Learning (ICML-10)}, 2010, pp. 399--406.

\bibitem{Rolfe2013DiscriminativeRS}
Jason~Tyler Rolfe and Yann LeCun,
\newblock ``Discriminative recurrent sparse auto-encoders,''
\newblock {\em CoRR}, vol. abs/1301.3775, 2013.

\bibitem{lewicki1998review}
Michael~S Lewicki,
\newblock ``A review of methods for spike sorting: the detection and
  classification of neural action potentials,''
\newblock {\em Network: Computation in Neural Systems}, vol. 9, no. 4, pp.
  R53--R78, 1998.

\bibitem{beck2009fast}
Amir Beck and Marc Teboulle,
\newblock ``A fast iterative shrinkage-thresholding algorithm for linear
  inverse problems,''
\newblock {\em SIAM journal on imaging sciences}, vol. 2, no. 1, pp. 183--202,
  2009.

\bibitem{Kingma2014AdamAM}
Diederik~P. Kingma and Jimmy Ba,
\newblock ``Adam: A method for stochastic optimization,''
\newblock {\em CoRR}, vol. abs/1412.6980, 2014.

\bibitem{Chen1998AtomicDB}
Scott~Saobing Chen, David~L. Donoho, and Michael~A. Saunders,
\newblock ``Atomic decomposition by basis pursuit,''
\newblock {\em SIAM Review}, vol. 43, pp. 129--159, 1998.

\bibitem{Smith2017CyclicalLR}
Leslie~N. Smith,
\newblock ``Cyclical learning rates for training neural networks,''
\newblock {\em 2017 IEEE Winter Conference on Applications of Computer Vision
  (WACV)}, pp. 464--472, 2017.

\end{thebibliography}

\end{document}